\documentclass[10pt,twocolumn,letterpaper]{article}

\usepackage{iccv}              

\usepackage{balance}

\usepackage{booktabs}
\usepackage{adjustbox}
\usepackage{pifont} 
\usepackage{xcolor} 
\usepackage{soul} 
\setulcolor{black} 
\setul{0.8ex}{0.6pt} 
\usepackage{caption}
\usepackage[numbers]{natbib}
\usepackage{mathtools} 
\usepackage{multirow}
\usepackage{multicol}
\usepackage{marvosym}
\usepackage{array}

\definecolor{iccvblue}{rgb}{0.21,0.49,0.74}
\usepackage[pagebackref,breaklinks,colorlinks,allcolors=iccvblue]{hyperref}

\def\paperID{14096} 
\def\confName{ICCV}
\def\confYear{2025}

\title{OmniDiff: A Comprehensive Benchmark for Fine-grained \\ Image Difference Captioning\vspace{-3.5mm}}

\author{Yuan Liu$^{1}$ \quad Saihui Hou$^{1}$\textsuperscript{\Letter} \quad Saijie Hou$^{2}$ \quad Jiabao Du$^{1}$\\
Shibei Meng$^{1}$ \quad Yongzhen Huang$^{1,3}$\\
$^{1}$School of Artificial Intelligence, Beijing Normal University\\
$^{2}$School of Artificial Intelligence, Beijing University of Posts and Telecommunications\\
$^{3}$WATRIX.AI\\
\href{https://yuan-liu-omnidiff.github.io}{https://yuan-liu-omnidiff.github.io}
}

\begin{document}
\twocolumn[{%
\renewcommand\twocolumn[1][]{#1}%
\maketitle
\vspace{-0.4in}
\begin{center}
    \centering
    \includegraphics[width=0.95\linewidth]{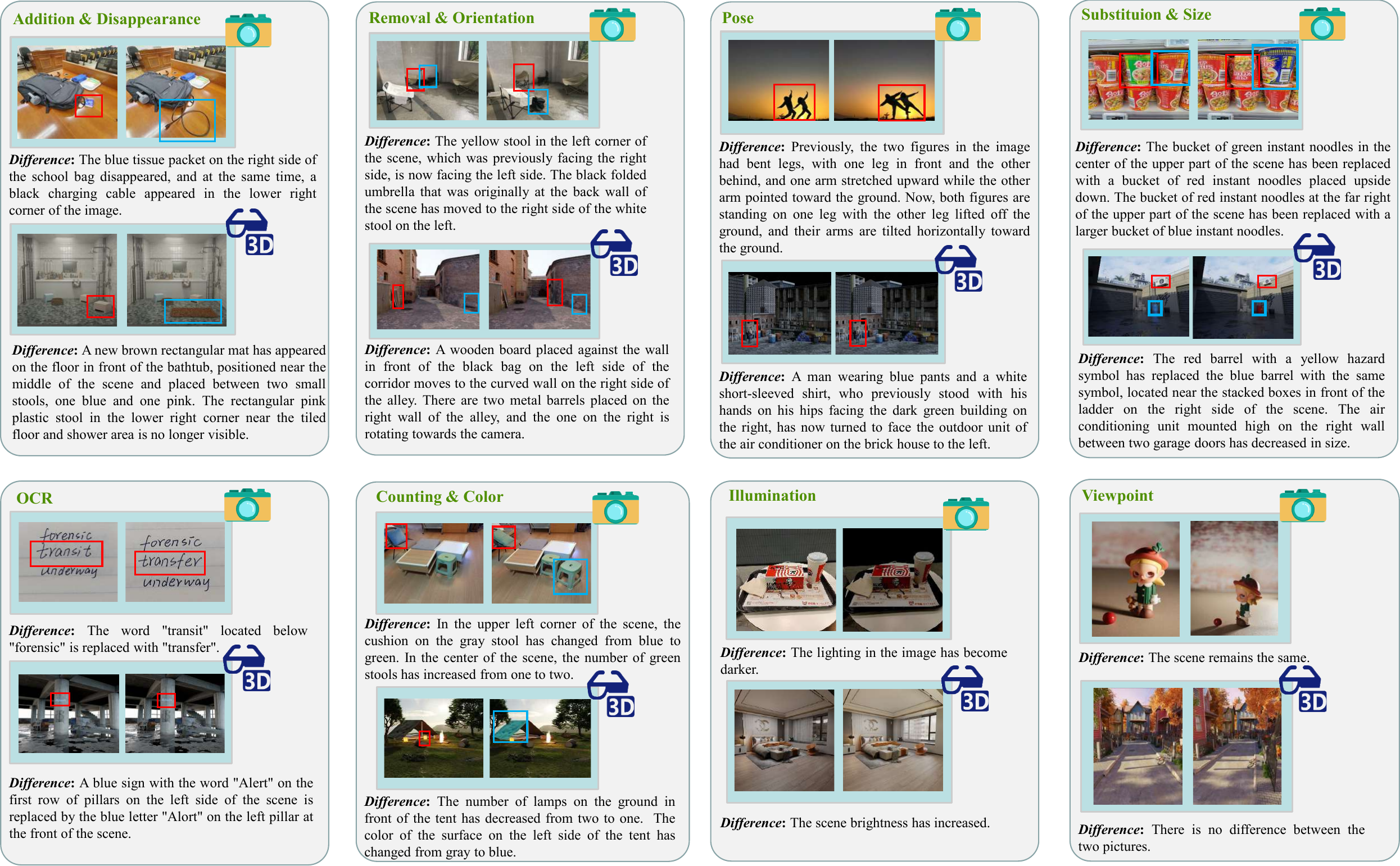}
    \captionof{figure}{
    OmniDiff: a comprehensive benchmark comprising 324 complex real-world and 3D synthetic environments, featuring 12 distinct types of variations, with an average difference caption length of 60 words per image pair.
    }
\label{fig:title}
\end{center}
}]
\begingroup 
\renewcommand\thefootnote{}
\footnotetext{\textsuperscript{\Letter} Corresponding author.} 
\endgroup 

\begin{abstract}
Image Difference Captioning (IDC) aims to generate natural language descriptions of subtle differences between image pairs, requiring both precise visual change localization and coherent semantic expression. Despite recent advancements, existing datasets often lack breadth and depth, limiting their applicability in complex and dynamic environments: (1) from a breadth perspective, current datasets are constrained to limited variations of objects in specific scenes, and (2) from a depth perspective, prior benchmarks often provide overly simplistic descriptions. To address these challenges, we introduce \textbf{OmniDiff}, a comprehensive dataset comprising 324 diverse scenarios—spanning real-world complex environments and 3D synthetic settings—with fine-grained human annotations averaging 60 words in length and covering 12 distinct change types. Building on this foundation, we propose \textbf{M$^3$Diff}, a \textbf{M}ulti\textbf{M}odal large language model enhanced by a plug-and-play \textbf{M}ulti-scale \textbf{Diff}erential Perception (MDP) module. This module improves the model's ability to accurately identify and describe inter-image differences while maintaining the foundational model's generalization capabilities. With the addition of the OmniDiff dataset, M$^3$Diff achieves state-of-the-art performance across multiple benchmarks, including Spot-the-Diff, IEdit, CLEVR-Change, CLEVR-DC, and OmniDiff, demonstrating significant improvements in cross-scenario difference recognition accuracy compared to existing methods. The dataset, code, and models will be made publicly available to support further research.
\vspace{-0.5cm}
\end{abstract}

\section{Introduction}
\label{sec:intro}

Image Difference Captioning (IDC) 
 focuses on describing subtle differences between two similar images. Unlike traditional change detection or single-image captioning, IDC requires both precise localization of change regions and accurate semantic expression of these changes, making it a more complex challenge in vision-language understanding. By bridging visual change detection and natural language generation, IDC has significant potential for applications in environmental monitoring and surveillance systems, where articulating visual differences is critical.

The emergence of domain-specific datasets and the refinement of vision-language models significantly enhance deep-learning-driven IDC methods. Despite notable progress in this field, limitations of existing datasets constrain the applicability of IDC methods in complex and dynamic environments. (1) From a \textbf{breadth} perspective, existing IDC datasets primarily focus on limited variations of objects in specific scenes, failing to cover the diverse changes encountered in real-world environments. For instance, Spot-the-Diff~\cite{jhamtani2018learning} only covers street surveillance footage from a fixed viewpoint, Birds-to-Words~\cite{forbes2019neural} focuses on fine-grained differences among similar bird species, and CLEVR-Change~\cite{park2019robust} renders simple desktop scenarios featuring five types of object changes. (2) Prior research often lacks \textbf{in-depth} discrepancy descriptions, which significantly restricts the model's ability to align visual differences with precise textual representations. For example, the average difference description between image pairs in IEdit~\cite{tan2019expressing} comprises merely 8 words, indicating that the captured changes are overly simplistic and fail to reflect the complexity of real-world variations.

To address these challenges, as illustrated in Figure~\ref{fig:title}, we introduce \textbf{OmniDiff}, a high-quality dataset comprising 324 complex real-world and 3D scenes, encompassing 12 distinct types of variations, each annotated with fine-grained human-labeled descriptions. The annotations exhibit an average length of 60 words, ensuring comprehensive and detailed representations of the observed variations. We collect diverse real-world scenes through on-site photography and web crawling, and further expand the dataset by constructing complex 3D scenes using Blender~\cite{blender} to simulate real-world variations. Unlike CLEVR-series~\cite{park2019robust,kim2021agnostic,qiu2021describing} datasets generated with Blender~\cite{blender}, which are confined to simplistic tabletop environments, our approach focuses on creating near-realistic complex settings, effectively enriching the data samples while simultaneously imposing higher demands on the model's 3D spatial perception capabilities.


Leveraging their foundational knowledge and strong generalization capabilities, Multimodal Large Language Models (MLLMs) demonstrate exceptional performance across a wide range of cross-modal tasks, including single-image captioning and visual question answering. This emerging trend motivates the exploration of applying MLLMs to the IDC task, which requires enhanced capabilities in \emph{fine-grained visual perception} and \emph{comprehensive multi-image content understanding}. 

In this work, to enhance the fine-grained difference perception capabilities of MLLMs between image pairs, we integrate a plug-and-play \textbf{M}ulti-scale \textbf{Diff}erential Perception (MDP) Module into the \textbf{M}ulti\textbf{M}odal Large Language Model framework, establishing \textbf{M$^3$Diff}, a strong base model designed for fine-grained IDC. Through our simple yet effective fine-tuning strategy, M$^3$Diff achieves state-of-the-art performance across a variety of benchmarks, including OmniDiff, Spot-the-Diff~\cite{jhamtani2018learning}, IEdit~\cite{tan2019expressing}, CLEVR-Change~\cite{park2019robust} and CLEVR-DC~\cite{kim2021agnostic}. These results demonstrate M$^3$Diff's strong generalization capabilities in recognizing differences across varied scenarios, highlighting its adaptability and effectiveness in addressing complex visual understanding tasks.


The primary contributions of this work are summarized as follows:
\begin{itemize}
    \item We construct \textbf{OmniDiff}, a high-quality dataset comprising 324 diverse scenarios, encompassing both real-world complex environments and 3D synthetic settings. The dataset is characterized by fine-grained human annotations, with an average description length of 60 words, and comprehensively covers 12 distinct types of changes.
    \item We integrate a plug-and-play Multi-scale Differential Perception (MDP) Module into the MLLM architecture to enhance fine-grained difference perception and multi-image content understanding capabilities. This approach facilitates the development of a strong base model, \textbf{M$^3$Diff}, specifically tailored for IDC.
    \item By leveraging the OmniDiff dataset and the M$^3$Diff architecture, we achieve state-of-the-art performance across multiple benchmarks spanning diverse scenarios.
\end{itemize}

\section{Related Works}

\subsection{Image Difference Captioning}
\noindent\textbf{Methods.} Research on IDC has evolved through several key methodological advancements. Jhamtani \etal~\cite{jhamtani2018learning} pioneer this task by aligning pixel-wise differences between image pairs. Subsequent works~\cite{park2019robust}~\cite{hosseinzadeh2021image}~\cite{liu2022remote} attempt to compute differences by directly subtracting image representations. However, these methods exhibit limited generalization capabilities for unaligned image pairs due to pseudo-changes caused by distractors (\eg, viewpoint, illumination). To address robustness against distractors, recent studies introduce specialized learning mechanisms. DIRL~\cite{tu2024distractors} stabilizes representations by correlating corresponding channels while decorrelating dissimilar ones, enabling robust difference extraction. Furthermore, multi-task learning frameworks integrate auxiliary objectives to enhance model accuracy. For example, Semantic-CC~\cite{zhu2024semantic} combines change detection (CD) and captioning (CC) via pixel-level semantic segmentation, improving linguistic precision. With the rise of vision-language pretraining, IDC-PCL~\cite{yao2022image} and CLIP4IDC~\cite{guo2022clip4idc} propose a two-stage pretrain-finetune framework to effectively model difference representations for change captioning. Recent MLLM-based approaches significantly advance IDC capabilities. FINER-MLLM~\cite{zhang2024differential} enhances change captioning through a LoRA~\cite{hu2021loralowrankadaptationlarge} fine-tuned MLLM with dual intra- and inter-image constraints, while Hu \etal~\cite{hu2024onediff} propose a generalist model combining a siamese encoder and Visual Delta Module for detecting and describing subtle differences.


\noindent\textbf{Benchmarks.} Benchmarks for change captioning are developed across various domains to address domain-specific challenges. For instance, Spot-the-Diff~\cite{jhamtani2018learning} and STVchrono~\cite{sun2024stvchrono} extract similar video frames from surveillance footage to capture changes in outdoor environments. Birds-to-Words~\cite{forbes2019neural} focuses on distinguishing subtle appearance differences among similar bird species, while LEVIR-CC~\cite{liu2022remote} targets changes in urban remote sensing image pairs. Unlike datasets tailored to specific scenes or objects, IEdit~\cite{tan2019expressing} comprises approximately 4,000 image pairs capturing diverse daily life scenarios. To address annotation scalability, CLEVR-Change~\cite{park2019robust} leverages the CLEVR engine~\cite{johnson2017clevr} to synthesize controlled variations in 3D-rendered tabletop scenes, with CLEVR-DC~\cite{kim2021agnostic} introducing extreme viewpoint-agnostic shifts and Qiu \etal~\cite{qiu2021describing} extending the framework through multi-change scenarios and unknown change counts for robust captioning evaluation.


\subsection{Multimodal Large Language Models}
Multimodal large language models exhibit exceptional capabilities in visual perception and natural language generation by aligning visual and textual features~\cite{alayrac2022flamingo,li2023blip} and leveraging visual instruction tuning~\cite{liu2023visual,liu2024improved,zhu2023minigpt,dai2023instructblipgeneralpurposevisionlanguagemodels}. Building on these foundations, recent advancements emphasize fine-grained visual understanding. Gromma~\cite{ma2024groma} bridges MLLMs and detection by embedding region-aware visual tokens directly into the language model’s latent space. Lai \etal~\cite{lai2024lisa} propose LISA, an MLLM that integrates segmentation via a token and embedding-as-mask mechanism for complex visual-textual reasoning segmentation. Pushing the boundaries of these advancements,~\cite{li2024llavanextinterleavetacklingmultiimagevideo,li2024llavaonevisioneasyvisualtask,bai2025qwen25vltechnicalreport,jiang2024mantis} further expand the scope of MLLMs by exploring their ability to understand multi-image and video content, demonstrating their potential for complex multimodal tasks. Despite these significant strides, applying MLLMs to IDC tasks presents unique challenges. IDC requires not only fine-grained visual perception but also the ability to precisely articulate subtle differences between images. Existing MLLMs often struggle to generate accurate and contextually rich descriptions of nuanced visual changes.
\section{The OmniDiff Dataset}
\label{sec:Dataset}

\subsection{Overview of OmniDiff}


We introduce \textbf{OmniDiff}, a high-quality dataset encompassing diverse real-world and synthetic complex scenarios, featuring fine-grained human annotations to ensure comprehensive and accurate labeling. OmniDiff comprises 324 scenes spanning diverse indoor and outdoor environments, collected through a combination of on-site photography, web scraping and 3D rendering. The dataset encompasses 12 categories of fine-grained visual variations, including \textit{viewpoint}, \textit{illumination}, \textit{addition}, \textit{disappearance}, \textit{removal}, \textit{substitution}, \textit{size}, \textit{color}, \textit{orientation}, \textit{pose}, \textit{OCR (textual changes)}, and \textit{counting}. Each similar image pair encompasses one or more types of variations, with the average length of difference descriptions extending to 60 words to ensure a thorough and detailed representation of the observed changes.




\begin{figure}[tp]
    \centering
    \includegraphics[width=\linewidth]{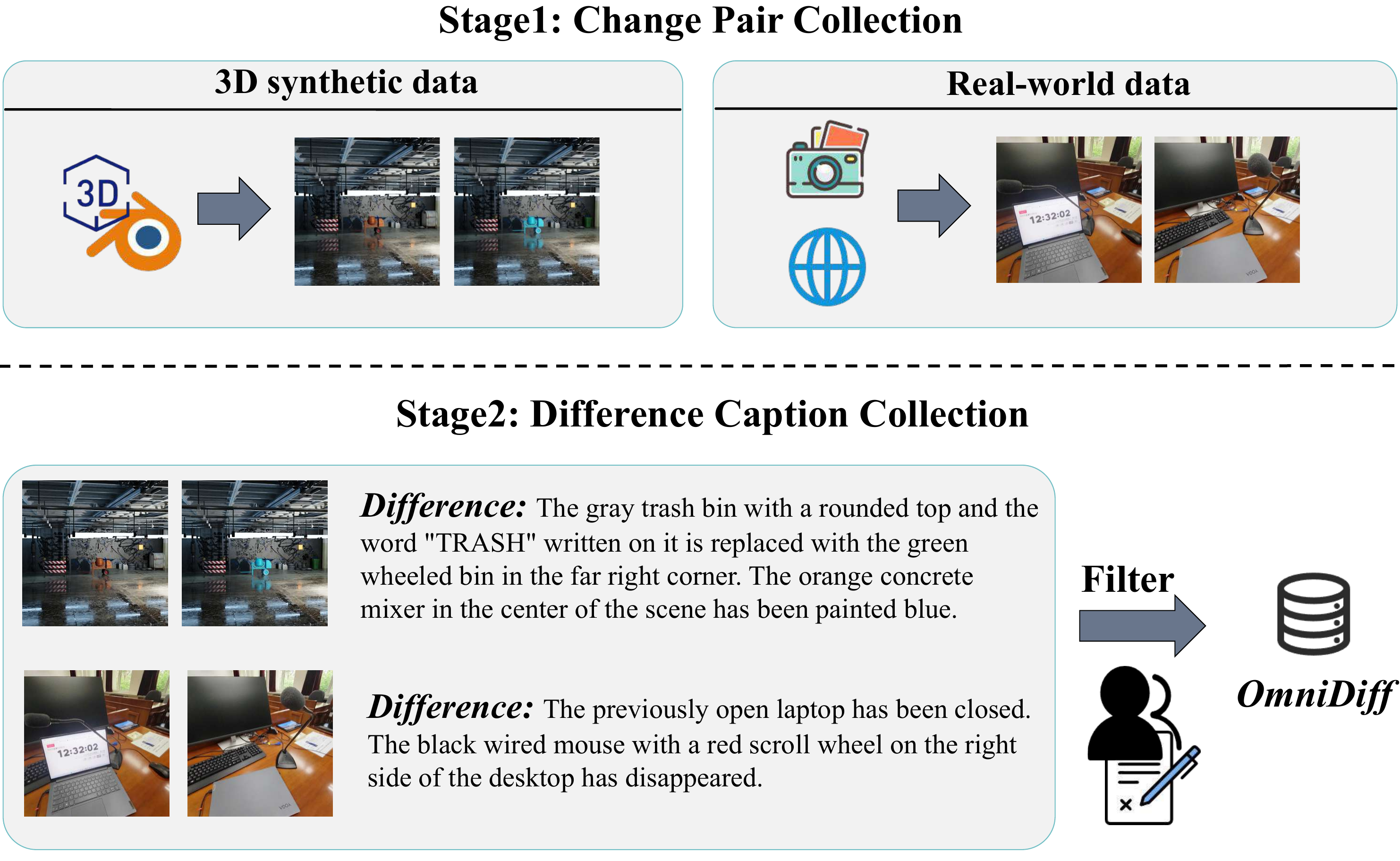}
    \captionof{figure}{
    Overview of the OmniDiff construction pipeline.
    }
\label{fig:data_pipeline}
\vspace{-0.4cm}
\end{figure}

\begin{table*}[tp]
\centering
\caption{The comparison of OmniDiff with existing image difference caption datasets. OmniDiff excels in both its breadth, encompassing a diverse range of real-world scenarios, and its depth, characterized by detailed and fine-grained annotations. The underlined value indicates a relatively high number of image pairs in CLEVR-Change~\cite{park2019robust}, although these pairs are limited to tabletop environments with constrained object variations. \textit{Distractor change} refers to non-semantic variations (\eg,viewpoint shifts and illumination changes).} 
\vspace{-0.2cm}
\renewcommand{\arraystretch}{1.1} 
\begin{adjustbox}{width=\textwidth}
\begin{tabular}{c|ccccccc}
\toprule
Dataset & Environment & Image source & Image pair & \begin{tabular}[c]{@{}c@{}}Distractor\\ change\end{tabular} & \begin{tabular}[c]{@{}c@{}}Human-labeled\\ caption\end{tabular} & \begin{tabular}[c]{@{}c@{}}Average words\\ per caption\end{tabular} & \begin{tabular}[c]{@{}c@{}}\% Long caption\\ (\textgreater 20 words)\end{tabular} \\ \midrule
CLEVR-Change~\cite{park2019robust} & table & 3D render &\underline{79606} & \ding{51} & \textcolor{lightgray}{\ding{55}} & 8 & 0\% \\
Spot-the-Diff~\cite{jhamtani2018learning} & outdoor & surveillance video & 13192 & \textcolor{lightgray}{\ding{55}} & \ding{51} & 19 & 19\% \\
LEVIR-CC~\cite{9934924} & outdoor & satellite & 10077 & \ding{51} & \ding{51} & 8 & 1\% \\
IEdit~\cite{tan2019expressing} & \textbf{in- \& out-door} & diverse real life & 3939 & \textcolor{lightgray}{\ding{55}} & \ding{51} & 8 & 2\% \\
Birds-to-Words~\cite{forbes2019neural} & outdoor & birds & 3347 & \textcolor{lightgray}{\ding{55}} & \ding{51} & 32 & 25\% \\ \midrule
\textbf{OmniDiff (ours)} & \textbf{in- \& out-door} & \textbf{3D render \& diverse real life} & \textbf{15598} & \ding{51} & \ding{51} & \textbf{60} & \textbf{37\%} \\ \bottomrule
\end{tabular}
\vspace{-0.4cm}
\end{adjustbox}
\label{OmniDiff_Compare}
\end{table*}


\subsection{Dataset Construction Process}
The construction process of the OmniDiff consists of two key stages: Change Pair Collection and Difference Caption Collection. Figure~\ref{fig:data_pipeline} shows the overall workflow.

\subsubsection{Change Pair Collection.} OmniDiff comprises change pairs sourced from 324 diverse indoor and outdoor scenes, encompassing a wide range of everyday environments such as streets, supermarkets, schools, and bedrooms, as illustrated in Figure \ref{fig:title}. The change pairs originate from two primary sources: real-world data and 3D synthetic data. 

To collect change pairs from real-world scenes, we utilize a combination of on-site photography and web crawling to capture diverse environmental conditions and various types of changes, ensuring comprehensive coverage of real-world scenarios. For on-site photography, we systematically position cameras in diverse indoor and outdoor environments to capture natural scene variations over time, while adjusting camera perspectives to simulate viewpoint changes and enrich the diversity of the collected data. To further enhance scene diversity, we collect online videos to capture environments and change types that are challenging to obtain through on-site photography. We specifically select videos depicting real-life scenarios and manually extract frames with similar scenes but subtle visual differences to construct change pairs. The sources and quality of these videos are rigorously screened to ensure compliance with the dataset's high-quality standards. Through a combination of on-site photography and web crawling, we collect change pairs from 224 distinct real scenes. 

In addition to real-world data, we utilize the Blender~\cite{blender} engine to render complex 3D scenes, simulating real-world variations. Unlike the CLEVR-series~\cite{park2019robust,kim2021agnostic,qiu2021describing}, which restricts changes to a limited set of objects within a single scene, we source high-quality 3D assets from ArtStation~\cite{artstation}, including 50 indoor and 50 outdoor scenes, and manually modify objects within the scenes to emulate realistic and diverse changes.

\subsubsection{Difference Caption Collection.} Although state-of-the-art MLLMs such as GPT-4o~\cite{gpt4} exhibit strong capabilities in generating textual descriptions, our experiments reveal their limitations in annotating fine-grained image differences in complex scenes, particularly in precisely localizing and describing subtle visual changes. Therefore, we rely on human annotators to ensure the accuracy and reliability of the dataset. Given that the images are captured in complex scenes, we instruct the annotators to accurately describe the location, attributes, and types of changes for the altered objects. To ensure consistency and quality, we provide the annotators with the following annotation guidelines: 1) Focus on describing semantic changes in the scene (\eg, the addition of an object) while ignoring irrelevant variations (\eg, viewpoint shifts). 2) Structure each caption into two parts: the \textit{referring part} (\eg, a man in a blue shirt standing behind a desk on the left side of the scene) and the \textit{change part} (\eg, changed from facing the window on the left to facing the sofa on the right). 

After completing the annotation process, we review and correct spelling and grammatical errors, resulting in a final collection of 15,598 difference captions with an average length of 60 words per caption.

\begin{table}[tbp]
\centering
\caption{Statistics of the OmniDiff Dataset.}
\vspace{-0.2cm}
\label{OmniDiff_stats}
\resizebox{\linewidth}{!}{ 
\begin{tabular}{lr}
\toprule
\textbf{Statistic} & \textbf{Value} \\
\midrule
Image Change Pairs & 15,598 \\
Total Captions & 15,598 \\
Avg Words per Caption & 60.0  \\  
Total Sentences & 38,653 \\
Sentences per Caption & 2.5 \\
Vocabulary Size & 3,793 \\
\midrule
Splits & \multicolumn{1}{l}{Train:Val.:Test = 80\%:10\%:10\%} \\
\bottomrule
\end{tabular}
}
\vspace{-0.3cm}
\end{table}
\subsection{Dataset Statistics}
As shown in Table~\ref{OmniDiff_stats}, OmniDiff contains a total of 15,598 change pairs, consisting of 8,609 pairs of real-world scene images and 6,989 pairs of 3D-rendered scene images, achieving a balanced ratio of 1.2:1 between real-world and synthetic data. The average difference description for each image pair consists of 60 words, ensuring a precise and detailed representation of fine-grained variations between images. We divide both indoor and outdoor scenes into training, validation, and test sets at an 8:1:1 ratio, ensuring that each subset comprehensively encompasses all 12 distinct types of changes.

\begin{figure*}[tp]
    \centering
    \includegraphics[width=0.8\linewidth]{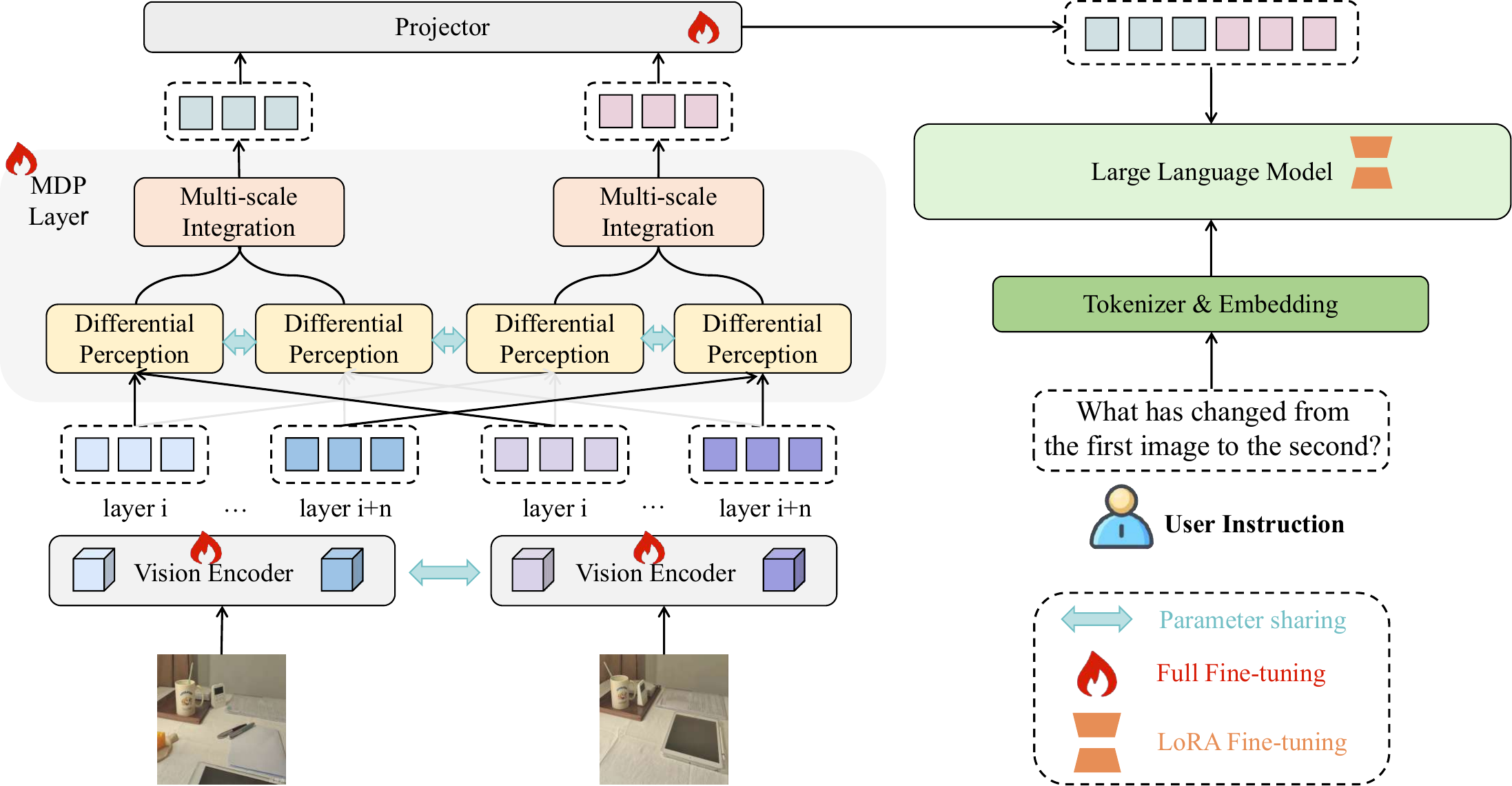}
    \captionof{figure}{
    The whole structure of M$^3$Diff with the multi-scale differential perception module.
    }
\label{fig:method}
\vspace{-0.2cm}
\end{figure*}

\subsection{Evaluation protocol}  For change captioning, we evaluate performance using five standard metrics: BLEU-4 (B)~\cite{Papineni_Roukos_Ward_Zhu_2001}, METEOR (M)~\cite{Banerjee_Lavie_2005}, ROUGE-L (R)~\cite{Lin_2004}, CIDEr-D (C)~\cite{Vedantam_Zitnick_Parikh_2015}, and SPICE (S)~\cite{Anderson_Fernando_Johnson_Gould_2016}. All metrics are computed using the Microsoft COCO evaluation server~\cite{chen2015microsoftcococaptionsdata}.

\subsection{Comparisons with Existing Datasets}
Table \ref{OmniDiff_Compare} further highlights the advantages of OmniDiff compared to other datasets. \textit{From the breadth perspective}, existing datasets primarily focus on limited variations of objects within single scenes (\eg, CLEVR-Change~\cite{park2019robust}), whereas OmniDiff combines real-world and 3D-rendered scenes to construct image pairs in complex and diverse environments, encompassing 12 distinct types of variations. \textit{In terms of depth}, previous datasets typically provide short annotations that lack the granularity needed to clearly describe fine-grained differences between image pairs. In contrast, OmniDiff includes a significant portion of long, detailed descriptions that precisely identify changes in complex scenes, offering a richer and more informative resource for training and evaluating multimodal models.

\section{Method}

\subsection{Overview} 
Inspired by recent advancements in leveraging MLLMs for change captioning tasks~\cite{wang2024ccexpert,zhu2024semantic}, our M$^3$Diff extends the standard MLLM architecture with a multi-scale differential perception module, as illustrated in Figure~\ref{fig:method}. This dedicated component implements feature-level differencing through channel-wise subtraction between image pairs, followed by adaptive cross-layer fusion operations that establish coherent difference representations prior to linguistic decoding. In this section, we first introduce the MDP module for multi-scale difference modeling, followed by a systematic exposition of the training paradigm of our framework.

\subsection{Multi-Scale Differential Perception} 
MDP module is employed to extract multi-scale discriminative discrepancy features through channel-wise subtraction, then fuses them with original image features via cross-attention mechanisms. By explicitly modeling cross-scale visual disparities and their contextual interactions, this module enhances precise difference localization and descriptive feature representation for captioning tasks.


\noindent \textbf{Differential Perception.}
The difference feature between image pairs is critical for IDC tasks but prone to distractors (\eg, illumination/viewpoint). The purpose of this operation is to generate a robust differential feature, and then integrate the enhanced difference representation back into the original feature streams for improved discriminability. 

Specifically, the differential feature is calculated by:
\begin{align}
\boldsymbol{\lambda}_1 = \sigma \big( \mathbf{W}_m & [\mathbf{F}_1^i \Vert \mathbf{F}_2^i] \big),  \ 
\boldsymbol{\lambda}_2 = \sigma \big( \mathbf{W}_m [\mathbf{F}_2^i \Vert \mathbf{F}_1^i] \big), \\
\hat{\mathbf{F}}_k^i & =  \mathbf{F}_k^i  \odot \boldsymbol{\lambda}_k,\ k\in\{1,2\}, \\
\Delta\mathbf{F}^i & = \mathbf{W}_p \big[ \hat{\mathbf{F}}_1^i \Vert \hat{\mathbf{F}}_2^i \Vert (\hat{\mathbf{F}}_1^i-\hat{\mathbf{F}}_2^i) \big],
\end{align}
where $\mathbf{F}_1^i,\mathbf{F}_2^i$ are image pair features from the $i$-th layer, $\Vert$ denotes concatenation, and $\sigma$ is sigmoid function. The weight $\mathbf{W}_m \in \mathbb{R}^{d\times 2d}$ processes concatenated features through a learnable projection, where $d$ denotes the feature dimension. The resulting attention weights $\boldsymbol{\lambda}_k \in [0,1]^d$ modulate input features via element-wise multiplication ($\odot$), preserving original dimensions. 
$\mathbf{W}_p \in \mathbb{R}^{d \times 3d}$ projects the concatenated triplet into $\Delta\mathbf{F}^i$, a $d$-dimensional vector that explicitly encodes feature variations.

Then, we fuse the salient differences with the original image features through cross-attention mechanisms:
\begin{align}
\Delta\mathbf{F}^{i}_\text{sa} &= \text{SA}(\Delta\mathbf{F}^i), \\
\Delta\mathbf{F}^{i}_\text{ca} &= \text{MLP} \Big( \text{CA} \big( \Delta\mathbf{F}^{i}_\text{sa},\ \mathbf{F}_{1}^{i} \Vert \mathbf{F}_{2}^{i} \Vert \Delta\mathbf{F}^{i}_\text{sa} \big) \Big), \\
\tilde{\mathbf{F}}^{i}_{k} &= \text{CA} \big( \mathbf{F}^{i}_{k},\ \mathbf{F}_{k}^{i}  \Vert \Delta\mathbf{F}^i_\text{ca} \big),\ k\in\{1,2\},
\end{align}
where the cross-attention (CA) establishes correspondences between two streams: (1) using the self-attention (SA) difference representation $\Delta\mathbf{F}^{i}_\text{sa}$ as queries, and (2) employing the concatenation of original features and $\Delta\mathbf{F}^{i}_\text{sa}$ as keys/values. After Multiple Layer Perceptron (MLP) processing, CA further fuses each original feature $\mathbf{F}^{i}_{k}$ with the refined difference signal $\mathbf{F}_{k}^{i}  \Vert \Delta\mathbf{F}^i_\text{ca}$ via query and key/value interactions, preserving spatial dimensions while enhancing channel-wise discriminability.

\begin{table*}[t]
\centering
\vspace{-0.2cm}
\caption{Performance comparison on OmniDiff across all metrics. The best results are highlighted in boldface, while the second-best results are indicated with underlining. \dag~indicates models that are trained without utilizing the OmniDiff dataset. }
\setlength{\tabcolsep}{5mm}
\resizebox{\textwidth}{!}{%
\begin{tabular}{>{\centering\arraybackslash}l|cccc|cccc}
\toprule
\multirow{2}{*}{Method} & \multicolumn{4}{c|}{Real} & \multicolumn{4}{c}{Render} \\ 
\cmidrule(lr){2-5} \cmidrule(lr){6-9}
 & BLEU-4 & METEOR & ROUGE-L & CIDEr & BLEU-4 & METEOR & ROUGE-L & CIDEr \\ \midrule
VARD-LSTM (TIP'23)~\cite{tu2023adaptive} & 5.5 & 12.8 & 24.2 & 6.7 & 4.2 & 10.2 & 24.7 & 3.7 \\
SCORER (ICCV'23)~\cite{tu2023self} & 7.2 & 11.1 & 23.5 & 7.0 & 9.9 & 13.3 & 27.1 & 4.1 \\
DIRL (ECCV'24)~\cite{tu2024distractors} & 7.5 & 11.0 & 23.8 & 5.4 & 11.7 & 13.8 & 27.6 & 5.9 \\
CARD (ACL'24)~\cite{tu2024contextawaredifferencedistillingmultichange} & \underline{9.1} & 12.6 & 25.1 & 9.2 & 11.3 & 13.4 & 27.3 & 7.3 \\ \midrule
\multicolumn{9}{c}{MLLM} \\ \midrule
\dag GPT-4o~\cite{gpt4} & 3.1 & 13.6 & 21.0 & 5.2 & 4.6 & 10.7 & 21.4 & 5.6 \\
\dag LLaVA-OneVision-7B~\cite{li2024llavaonevisioneasyvisualtask} & 0.2 & 4.1 & 13.6 & 1.7 & 0.3 & 4.7 & 15.6 & 1.6 \\
\dag Qwen-2.5-VL-7B~\cite{bai2025qwen25vltechnicalreport} & 3.8 & 9.5 & 19.8 & 6.2 & 2.1 & 6.8 & 18.3 & 3.3 \\ \midrule
FINER-MLLM (MM'24)~\cite{zhang2024differential} & 8.9 & \underline{13.8} & \underline{25.9} & \underline{11.7} & \underline{13.6} & \underline{15.6} & \underline{29.9} & \underline{14.0} \\
\textbf{M$^3$Diff (ours)} & \textbf{14.3} & \textbf{18.9} & \textbf{32.9} & \textbf{31.3} & \textbf{15.7} & \textbf{19.9} & \textbf{35.3} & \textbf{28.3}  \\ \bottomrule
\end{tabular}%
}
\vspace{-0.2cm}
\label{tab:omnidiff}
\end{table*}

\noindent \textbf{Multi-Scale Integration.}
MLLMs typically employ the visual features from the penultimate layer of the visual encoder~\cite{liu2023visual,liu2024improved,li2024llavaonevisioneasyvisualtask}.
However, features from individual network layers face inherent limitations: low-level features lack semantic consistency, while high-level features lose detail perception. Therefore, multi-layer feature fusion is employed to aggregate low-level details and high-level semantics to yield a more comprehensive and discriminative representation. The fusion process can be formulated as:
\begin{align}
\text{Score}^i &= \sigma \Big(\text{MLP}\big( \Phi_\text{Mean}( \tilde{\mathbf{F}}_1^i \Vert \tilde{\mathbf{F}}_2^i \Vert \Delta \mathbf{F}^i ) \big) \Big), \ \\
\mathbf{F}_k' &= \sum\nolimits_i  \text{Score}^i \odot \tilde{\mathbf{F}}_k^i, \ k\in\{1,2\},
\end{align}
where $\Phi_\text{Mean}$ stands for the token-wise mean pooling to compress spatial information. $\text{Score}^i$ is the fusion weight of the $i$-th layer's feature. 
The weights are then used to compute the refined features $\mathbf{F}_1'$ and $\mathbf{F}_2'$ via element-wise multiplication and cross-layer summation.

\subsection{Training Strategy} 


Unlike previous multi-stage approaches~\cite{zhu2024semantic,hu2024onediff,wang2024ccexpert} adapting MLLMs for IDC tasks, we employ a simple yet effective one-stage fine-tuning strategy. Despite a gap between the MDP module and the well-pretrained MLLM~\cite{li2024llavaonevisioneasyvisualtask}, large-scale instruction tuning enables the lightweight MDP module to enhance difference perception without disrupting existing knowledge, achieving plug-and-play functionality.

The dataset utilized for fine-tuning M$^3$Diff consists of 896k question-answer pairs  (refer to Section 8.1 of the Supplementary Material for details). In addition to OmniDiff, the dataset integrates several publicly available IDC datasets, including Spot-the-Diff~\cite{jhamtani2018learning}, IEdit~\cite{tan2019expressing}, and Birds-to-Words~\cite{forbes2019neural}, as well as 3D-rendered samples from CLEVR-Change~\cite{park2019robust} and CLEVR-DC~\cite{kim2021agnostic}.  During training, we apply LoRA~\cite{hu2021loralowrankadaptationlarge} for parameter-efficient fine-tuning to the LLM while fully fine-tuning the rest of the model.

\section{Experiments}

\subsection{Datasets}
We evaluate our model on four public datasets and OmniDiff, including the real-world domain datasets Spot-the-Diff~\cite{jhamtani2018learning}, Image-Editing-Request~\cite{tan2019expressing}, and OmniDiff-Real, as well as the 3D-rendered datasets CLEVR-Change~\cite{park2019robust}, CLEVR-DC~\cite{kim2021agnostic}, and OmniDiff-Render.

\textbf{Spot-the-Diff}~\cite{jhamtani2018learning} consists of 13,192 pairs of similar video frames captured from street surveillance footage. Following the standard evaluation protocol used in prior work~\cite{jhamtani2018learning}, we evaluate our model in a single-change captioning setting. The dataset is officially split into training, validation, and testing sets with a ratio of 8:1:1.

\textbf{Image-Edit-Request }~\cite{tan2019expressing} consists of 3,939 aligned image pairs from real-life scenarios, accompanied by 5,695 editing instructions. Following the official dataset split protocol~\cite{tan2019expressing}, we use 3,061 image pairs for training, 383 for validation, and 495 for testing.

\textbf{CLEVR-Change}~\cite{park2019robust} is a large-scale synthetic dataset featuring scene changes involving geometric objects and distractors under moderate viewpoint variations. The dataset contains 79,606 image pairs, split into 67,660 for training, 3,976 for validation, and 7,970 for testing.

\textbf{CLEVR-DC}~\cite{kim2021agnostic} is a large-scale dataset featuring extreme viewpoint shifts, comprising 48,000 image pairs with the same change types as CLEVR-Change~\cite{park2019robust}. The dataset is split into 85\% training, 5\% validation, and 10\% testing.

\textbf{OmniDiff-Real}, a subset of the OmniDiff dataset, comprises 8,609 real-world image pairs, split into 7,007 for training, 799 for validation, and 803 for testing. Given the diverse and complex changes in OmniDiff, our model is evaluated in a multi-change captioning setting.

\textbf{OmniDiff-Render} comprises 3D-rendered image pairs from the OmniDiff dataset, with 5,445 pairs for training, 777 for validation, and 767 for testing. Model evaluation is performed under the multi-change setting, consistent with the approach used for the OmniDiff-Real.

\subsection{Implementation Details}
\noindent\textbf{Architecture.}
M$^3$Diff initializes the weights of the visual encoder, projector, and LLM from LLaVA-OneVision-7B~\cite{li2024llavaonevisioneasyvisualtask}, while the MDP module weights are newly initialized. The visual encoder utilizes the SigLIP model~\cite{zhai2023sigmoidlosslanguageimage}, containing 27 transformer layers. Multi-scale features extracted from layers 17, 20, 23, and 26 of the visual encoder serve as input to MDP, which comprises two stacked transformer layers. The projector comprises a two-layer multilayer perceptron, and the LLM is based on the Qwen2~\cite{yang2024qwen2}.

\begin{table}[tp]
\centering
\setlength{\tabcolsep}{1mm} %
\renewcommand\arraystretch{1.1}
\caption{Performance comparison on Spot-the-Diff across all metrics. The best results are highlighted in boldface, while the second-best results are indicated with underlining.}
\vspace{-0.2cm}
\resizebox{\linewidth}{!}{
\begin{tabular}{lccccc}
\toprule
\multicolumn{1}{l|}{Method} & BLEU-4 & METEOR & ROUGE-L & CIDEr & SPICE \\ \midrule
\multicolumn{1}{l|}{DUDA (ICCV'19)~\cite{park2019robust}} & 8.1 & 11.8 & 29.1 & 32.5 & - \\
\multicolumn{1}{l|}{M-VAM (ECCV'20)~\cite{shi2020findingsideviewpointadaptedmatching}} & 10.1 & 12.4 & 31.3 & 38.1 & - \\
\multicolumn{1}{l|}{M-VAM+RAF (ECCV'20)~\cite{shi2020findingsideviewpointadaptedmatching}} & 11.1 & 12.9 & 33.2 & 42.5 & 17.1 \\
\multicolumn{1}{l|}{IFDC (TMM'22)~\cite{huang2021image}} & 8.7 & 11.7 & 30.2 & 37.0 & - \\
\multicolumn{1}{l|}{DUDA+TIRG (CVPR'21)~\cite{hosseinzadeh2021image}} & 8.1 & 12.5 & 29.9 & 34.5 & - \\
\multicolumn{1}{l|}{VACC (ICCV'21)~\cite{kim2021agnostic}} & 9.7 & 12.6 & 32.1 & 41.5 & - \\
\multicolumn{1}{l|}{SRDRL (ACL'21)~\cite{tu2021semantic}} & - & 13.0 & 31.0 & 35.3 & 18.0 \\
\multicolumn{1}{l|}{R$^3$Net (EMNLP'21)~\cite{tu2021r}} & - & 13.1 & 32.6 & 36.6 & 18.8 \\
\multicolumn{1}{l|}{MCCFormers-D (ICCV'21)~\cite{qiu2021describing}} & 10.0 & 12.4 & - & 43.1 &  18.3  \\
\multicolumn{1}{l|}{CLIP4IDC (AACL'22)~\cite{guo2022clip4idc}} & 11.6 & 14.2 & 35.0 & 47.4 & - \\
\multicolumn{1}{l|}{VARD-Trans (TIP'23)~\cite{tu2023adaptive}} & - & 12.5 & 29.3 & 30.3 & 17.3 \\
\multicolumn{1}{l|}{SCORER (ICCV'23)~\cite{tu2023self}} & 10.2 & 12.2 & - & 38.9 & 18.4 \\
\multicolumn{1}{l|}{SMARL (TPAMI'24)~\cite{tu2024smart}} & - & 13.1 & 32.8 & 40.0 & 19.5 \\
\multicolumn{1}{l|}{SMART (TPAMI'24)~\cite{tu2024smart}} & - & 13.5 & 31.6 & 39.4 & 19.0 \\
\multicolumn{1}{l|}{DIRL (ECCV'24)~\cite{tu2024distractors}} & 10.3 & 13.8 & 32.8 & 40.9 & 19.9 \\ \midrule
\multicolumn{6}{c}{MLLM} \\ \midrule
\multicolumn{1}{l|}{FINER-MLLM (MM'24)~\cite{zhang2024differential}} & \underline{12.9} & \underline{14.7} & 35.5 & \underline{61.8} & \underline{22.1} \\
\multicolumn{1}{l|}{OneDiff (ACCV'24)~\cite{hu2024onediff}} & 12.8 & 14.6 & \underline{35.8} & 56.6 & - \\ \midrule
\multicolumn{1}{l|}{\textbf{M$^3$Diff (ours)}} & \textbf{14.4} & \textbf{15.4} & \textbf{37.6} & \textbf{71.1} & \textbf{24.9} \\ \bottomrule
\end{tabular}
}
\vspace{-0.4cm}
\label{tab:spot}
\end{table}

\noindent\textbf{Training setting.} 
Building upon the OmniDiff, we collect and curate publicly available real-world datasets, including Spot-the-Diff~\cite{jhamtani2018learning}, IEdit~\cite{tan2019expressing}, and a subset of Birds-to-Words~\cite{forbes2019neural}, alongside 3D-rendered datasets such as CLEVR-Change~\cite{park2019robust} and CLEVR-DC~\cite{kim2021agnostic}. These datasets are transformed into a question-answer format, yielding a fine-tuning dataset of 896K samples. For the CLEVR-Change~\cite{park2019robust} and CLEVR-DC~\cite{kim2021agnostic}, we utilize each caption per image pair as an independent training sample. During the fine-tuning phase, we employ LoRA~\cite{hu2021loralowrankadaptationlarge} for parameter-efficient tuning of the LLM to minimize computational costs while preserving the LLM's inherent knowledge. The LoRA configuration includes a rank of 128 and an alpha value of 256. Concurrently, the visual encoder, projector, and MDP module undergo full-parameter fine-tuning. The AdamW optimizer sets the initial learning rate to 2e-6 for the visual encoder and 1e-5 for the remaining modules. During training, a cosine annealing scheduler with a linear warm-up dynamically adjusts the learning rate. Our experiments utilize 8 NVIDIA A100 (40G) GPUs with a global batch size of 256, completing in 26 hours.

\begin{table}[tp]
\centering
\setlength{\tabcolsep}{1mm}
\renewcommand\arraystretch{1.1}
\caption{Performance comparison on Image-Edit-Request across all metrics. The best results are highlighted in boldface, while the second-best results are indicated with underlining.}
\vspace{-0.2cm}
\resizebox{\linewidth}{!}{
\begin{tabular}{lccccc}
\toprule
\multicolumn{1}{l|}{Method} & BLEU-4 & METEOR & ROUGE-L & CIDEr & SPICE \\ \midrule
\multicolumn{1}{l|}{DUDA (ICCV'19)~\cite{park2019robust}} & 6.5 & 12.4 & 37.3 & 22.8 & - \\
\multicolumn{1}{l|}{MCCFormers-D (ICCV'21)~\cite{qiu2021describing}} & 8.3 & 14.3 & 39.2 & 30.2 & - \\
\multicolumn{1}{l|}{CLIP4IDC (AACL'22)~\cite{guo2022clip4idc}} & 8.2 & 14.6 & 40.4 & 32.2 & - \\
\multicolumn{1}{l|}{NCT (TMM'23)~\cite{tu2023neighborhoodcontrastivetransformerchange}} & 8.1 & 15.0 & 38.8 & 34.2 & 12.7 \\
\multicolumn{1}{l|}{VARD-Trans (TIP'23)~\cite{tu2023adaptive}} & 10.0 & 14.8 & 39.0 & 35.7 & - \\
\multicolumn{1}{l|}{SCORER (ICCV'23)~\cite{tu2023self}} & 10.0 & 15.0 & 39.6 & 33.4 & - \\
\multicolumn{1}{l|}{SMARL (TPAMI'24)~\cite{tu2024smart}} & 10.4 & 15.1 & 40.3 & 34.6 & - \\
\multicolumn{1}{l|}{SMART (TPAMI'24)~\cite{tu2024smart}} & 10.5 & 15.2 & 39.1 & 37.8 & - \\
\multicolumn{1}{l|}{DIRL (ECCV'24)~\cite{tu2024distractors}} & 10.9 & 15.0 & 41.0 & 34.1 & - \\ \midrule
\multicolumn{6}{c}{MLLM} \\ \midrule
\multicolumn{1}{l|}{FINER-MLLM (MM'24)~\cite{zhang2024differential}} & 14.1 & 15.9 & 40.4 & 53.5 & \underline{15.9} \\
\multicolumn{1}{l|}{OneDiff (ACCV'24)~\cite{hu2024onediff}} & \underline{29.6} & \underline{25.1} & \underline{55.6} & \underline{109.6} & - \\ \midrule
\multicolumn{1}{l|}{\textbf{M$^3$Diff (ours)}} & \textbf{33.6} & \textbf{26.5} & \textbf{59.7} & \textbf{136.6} & \textbf{27.5} \\ \bottomrule
\end{tabular}
}
\vspace{-0.4cm}
\label{iedit}
\end{table}

\begin{table*}[t]
\centering
\renewcommand\arraystretch{0.9}
\vspace{-0.2cm}
\caption{Ablation Study of the OmniDiff dataset and MDP Module on OmniDiff and IEdit Benchmarks. }
\setlength{\tabcolsep}{3.5mm}
\resizebox{\textwidth}{!}{ 
\begin{tabular}{l|cccc|cccc|ccccc}
\toprule
\multirow{2}{*}{Settings} & \multicolumn{4}{c|}{OmniDiff-Real} & \multicolumn{4}{c|}{OmniDiff-Render} & \multicolumn{5}{c}{IEdit} \\
\cmidrule(lr){2-5} \cmidrule(lr){6-9} \cmidrule(lr){10-14}
 & B & M & R & C & B & M & R & C & B & M & R & C & S \\
\midrule
w/o OmniDiff \& MDP     & 0.1 & 3.3 & 11.5 & 1.1 & 0.0 & 2.9 & 10.0 & 0.7 & 31.5 & 26.0 & 59.6 & 133.5 & 25.7 \\
w/o OmniDiff                & 0.1 & 3.8 & 12.7 & 1.9 & 0.1 & 3.7 & 11.9 & 1.1 & 32.3 & 26.3 & 59.7 & 132.8 & 27.0 \\
w/o MDP           & 12.2 & 17.1 & 32.3 & \textbf{35.3} & 15.5 & 18.6 & 34.3 & 25.4 & 33.5 & 26.3 & 59.7 & 135.2 & 26.5 \\
\textbf{M$^3$Diff (ours)}  & \textbf{14.3} & \textbf{18.9} & \textbf{32.9} & 31.3 & \textbf{15.7} & \textbf{19.9} & \textbf{35.3} & \textbf{28.3} & \textbf{33.6} & \textbf{26.5} & \textbf{59.7} & \textbf{136.6} & \textbf{27.5} \\
\bottomrule
\end{tabular}
}
\label{merged-ablation}
\vspace{-0.1cm}
\end{table*}

\subsection{Performance Comparison}
To validate the effectiveness of M$^3$Diff in complex and dynamic scenarios, we fine-tune state-of-the-art IDC methods (VARD~\cite{tu2023adaptive}, SCORER~\cite{tu2023self}, DIRL~\cite{tu2024distractors}, CARD~\cite{tu2024contextawaredifferencedistillingmultichange}) and the MLLM-based approach FINER-MLLM~\cite{zhang2024differential} on the OmniDiff dataset. We then conduct a comprehensive evaluation on the test set to compare their performance. As demonstrated in Table \ref{tab:omnidiff}, M$^3$Diff achieves superior performance across all metrics in both real-world and rendered scenarios. Additionally, we perform zero-shot evaluations of leading MLLMs (GPT-4o~\cite{gpt4}, LLaVA-OneVision~\cite{li2024llavaonevisioneasyvisualtask}, and Qwen-2.5-VL~\cite{bai2025qwen25vltechnicalreport}) on OmniDiff, demonstrating their limitations in analyzing complex scene differences despite their general multimodal capabilities. 

As shown in Tables~\ref{tab:spot}, \ref{iedit}, \ref{tab:clevr-change}, and~\ref{tab:clevr-dc}, M$^3$Diff consistently achieves SOTA performance across various publicly IDC benchmarks, including real-world scenarios such as Spot-the-Diff~\cite{jhamtani2018learning} and IEdit~\cite{tan2019expressing}, as well as rendered environments like CLEVR-Change~\cite{park2019robust} and CLEVR-DC~\cite{kim2021agnostic}. These results demonstrate the model's strong generalization ability in cross-scenario difference identification.

\begin{table}[tp]
\centering
\setlength{\tabcolsep}{1mm}
\renewcommand\arraystretch{1.1}
\vspace{-0.2cm}
\caption{Performance comparison on CLEVR-Change across all metrics. The best results are highlighted in boldface, while the second-best results are indicated with underlining.}
\resizebox{\linewidth}{!}{
    \begin{tabular}{cccccc}
    \toprule
    \multicolumn{1}{l|}{Method} & BLEU-4 & METEOR & ROUGE-L & CIDEr & SPICE \\ \midrule
    \multicolumn{1}{l|}{DUDA (ICCV'19)~\cite{park2019robust}} & 47.3 & 33.9 & - & 112.3 & 24.5 \\
    \multicolumn{1}{l|}{M-VAM (ECCV'20)~\cite{shi2020findingsideviewpointadaptedmatching}} & 50.3 & 37.0 & 69.7 & 114.9 & 30.5 \\
    \multicolumn{1}{l|}{M-VAM+RAF (ECCV'20)~\cite{shi2020findingsideviewpointadaptedmatching}} & 51.3 & 37.8 & 70.4 & 115.8 & 30.7 \\
    \multicolumn{1}{l|}{IFDC (TMM'22)~\cite{huang2021image}} & 49.2 & 32.5 & 69.1 & 118.7 & - \\
    \multicolumn{1}{l|}{DUDA+TIRG (CVPR'21)~\cite{hosseinzadeh2021image}} & 51.2 & 37.7 & 70.5 & 115.4 & 31.1 \\
    \multicolumn{1}{l|}{VACC (ICCV'21)~\cite{kim2021agnostic}} & 52.4 & 37.5 & - & 114.2 & 31.0 \\
    \multicolumn{1}{l|}{SRDRL (ACL'21)~\cite{tu2021semantic}} & 54.9 & 40.2 & 73.3 & 122.2 & 32.9 \\
    \multicolumn{1}{l|}{R$^3$Net (EMNLP'21)~\cite{tu2021r}
    } & 54.7 & 39.8 & 73.1 & 123.0 & 32.6 \\
    \multicolumn{1}{l|}{MCCFormers-D (ICCV'21)~\cite{qiu2021describing}} & 52.4 & 38.3 & - & 121.6 & 26.8   \\
    \multicolumn{1}{l|}{CLIP4IDC (AACL'22)~\cite{guo2022clip4idc}} & \underline{56.9} & 38.4 & \textbf{76.4} & \textbf{150.7} & - \\
    \multicolumn{1}{l|}{NCT (TMM'23)~\cite{tu2023neighborhoodcontrastivetransformerchange}} & 55.1 & 40.2 & 73.8 & 124.1 & 32.9 \\
    \multicolumn{1}{l|}{VARD-Trans (TIP'23)~\cite{tu2023adaptive}} & 55.4 & 40.1 & 73.8 & 126.4 & 32.6 \\
    \multicolumn{1}{l|}{SCORER (ICCV'23)~\cite{tu2023self}} & 56.3 & \underline{41.2} & 74.5 & 126.8 & 33.3 \\
    \multicolumn{1}{l|}{SMARL (TPAMI'24)~\cite{tu2024smart}} & 55.7 & 40.6 & 73.8 & 126.5 & 33.4 \\
    \multicolumn{1}{l|}{SMART (TPAMI'24)~\cite{tu2024smart}} & 56.1 & 40.8 & 74.2 & 127.0 & \underline{33.4} \\
    \multicolumn{1}{l|}{DIRL (ECCV'24)~\cite{tu2024distractors}} & 54.6 & 38.1 & 71.9 & 123.6 & 31.8 \\ \midrule
    \multicolumn{6}{c}{MLLM} \\ \midrule
    \multicolumn{1}{l|}{FINER-MLLM (MM'24)~\cite{zhang2024differential}} & 55.6 & 36.6 & 72.5 & \underline{137.2} & 26.4 \\ \midrule
    \multicolumn{1}{l|}{\textbf{M$^3$Diff (ours)}} & \textbf{57.1} & \textbf{41.9} & \underline{75.3} & 130.5 & \textbf{33.8} \\ \bottomrule
    \end{tabular}
}
\vspace{-0.2cm}
\label{tab:clevr-change}
\end{table}


\subsection{Ablation Study and Analysis}

\noindent\textbf{Effects of OmniDiff Dataset.} 
Table~\ref{merged-ablation} illustrates the impact of fine-tuning M$^3$Diff with the OmniDiff dataset, demonstrating its performance improvements on both the OmniDiff benchmark and IEdit~\cite{tan2019expressing}. The experiments reveal that fine-tuning M$^3$Diff solely on existing public datasets fails to effectively describe differences between image pairs in OmniDiff, which contains a wide range of complex and dynamic scenarios. By incorporating the OmniDiff dataset into the fine-tuning process alongside existing public datasets, the model not only achieves significant performance improvements in complex scenarios within the OmniDiff benchmark but also demonstrates enhanced capabilities on IEdit~\cite{tan2019expressing}. These results validate that the diverse and dynamic samples in OmniDiff effectively strengthen the generalization ability of M$^3$Diff across varied scenarios.


\begin{table}[tp]
\centering
\setlength{\tabcolsep}{1mm}
\renewcommand\arraystretch{1.1}
\vspace{-0.2cm}
\caption{Performance comparison on CLEVR-DC across all metrics. The best results are highlighted in boldface, while the second-best results are indicated with underlining.}
\resizebox{\linewidth}{!}{
\begin{tabular}{lccccc}
\toprule
\multicolumn{1}{l|}{Method} & BLEU-4 & METEOR & ROUGE-L & CIDEr & SPICE \\ \midrule
\multicolumn{1}{l|}{DUDA (ICCV'19)~\cite{park2019robust}} & 40.3 & 27.1 & - & 56.7 & 16.1 \\
\multicolumn{1}{l|}{DUDA+CC (ICCV'19)~\cite{park2019robust}} & 41.7 & 27.5 & - & 62.0 & 16.4 \\
\multicolumn{1}{l|}{M-VAM (ECCV'20)~\cite{shi2020findingsideviewpointadaptedmatching}} & 40.9 & 27.1 & - & 60.1 & 15.8 \\
\multicolumn{1}{l|}{M-VAM+CC (ECCV'20)~\cite{shi2020findingsideviewpointadaptedmatching}} & 41.0 & 27.2 & - & 62.0 & 15.7 \\
\multicolumn{1}{l|}{VA (ICCV'21)~\cite{kim2021agnostic}} & 44.5 & 29.2 & - & 70.0 & 17.1 \\
\multicolumn{1}{l|}{VACC (ICCV'21)~\cite{kim2021agnostic}} & 45.0 & 29.3 & - & 71.7 & \underline{17.6} \\
\multicolumn{1}{l|}{MCCFormers-D (ICCV'21)~\cite{qiu2021describing}} & 46.9 & 31.7 & - & 71.6 & 14.6   \\
\multicolumn{1}{l|}{NCT (TMM'23)~\cite{tu2023neighborhoodcontrastivetransformerchange}} & 47.5 & 32.5 & 65.1 & 76.9 & 15.6 \\
\multicolumn{1}{l|}{VARD-Trans (TIP'23)~\cite{tu2023adaptive}} & 48.3 & 32.4 & - & 77.6 & 15.4 \\
\multicolumn{1}{l|}{SCORER (ICCV'23)~\cite{tu2023self}} & 49.4 & \underline{33.4} & 66.1 & 83.7 & 16.2 \\
\multicolumn{1}{l|}{DIRL (ECCV'24)~\cite{tu2024distractors}} & \underline{51.4} & 32.3 & \underline{66.3} & \underline{84.1} & 16.8 \\ \midrule
\multicolumn{6}{c}{MLLM} \\ \midrule
\multicolumn{1}{l|}{\textbf{M$^3$Diff (ours)}} & \textbf{60.6} & \textbf{37.6} & \textbf{73.0} & \textbf{109.4} & \textbf{21.3} \\ \bottomrule
\end{tabular}
}
\vspace{-0.4cm}
\label{tab:clevr-dc}
\end{table}

\noindent\textbf{Effects of Multi-Scale Differential Perception Module.} 
Table~\ref{merged-ablation} demonstrates that introducing the MDP module to enhance difference perception capabilities, while preserving the inherent knowledge of MLLM, leads to further improvements on both the OmniDiff and IEdit~\cite{tan2019expressing} test set. 
Removing MDP leads to consistent degradation in text quality and semantic coherence, highlighting its role in harmonizing multi-scale feature integration. While partial metric fluctuations occur (\ie, the CIDEr variation on OmniDiff-Real), the full model demonstrates balanced superiority, outperforming ablated versions in 12/13 metrics. This holistic enhancement confirms MDP’s necessity for fine-grained IDC tasks.

\section{Conclusion and Future Outlook}
In this work, we tackle the limitations of IDC by introducing \textbf{OmniDiff}, a comprehensive dataset featuring 324 diverse scenarios spanning real-world and 3D synthetic environments. With fine-grained human annotations averaging 60 words and covering 12 distinct change types, OmniDiff sets a new benchmark for breadth and depth in IDC datasets. Building on this, we propose \textbf{M$^3$Diff}, a MultiModal Large Language Model enhanced by a plug-and-play Multi-scale Differential Perception (MDP) module. Extensive experiments show that M$^3$Diff achieves state-of-the-art performance on multiple benchmarks, demonstrating significant improvements in cross-scenario difference recognition. 

In future work, OmniDiff can be expanded to include image sets capturing continuous real-world changes, enabling more comprehensive modeling of dynamic scenarios. Building on this, M$^3$Diff can further be extended to video-based difference captioning, significantly broadening its applicability to temporally evolving environments.

\section*{Acknowledgement}
This work is jointly supported by National Natural Science Foundation of China (62206022, 62276025, 62476027) and the Fundamental Research Funds for the Central Universities (2253200026).





{
    \small
    \bibliographystyle{ieeenat_fullname}
    \bibliography{main}
}

\setcounter{figure}{3}
\setcounter{table}{8}
\setcounter{section}{6}

\clearpage

\maketitlesupplementary
In this supplementary material, we provide additional details regarding the \textbf{OmniDiff} dataset and the training specifics of the \textbf{M$^3$Diff} model. 
Finally, we further experimentally validate the effectiveness of the OmniDiff dataset and the MDP module, and analyze the qualitative performance of four models on samples from OmniDiff.

\section{Dataset Statistics}
\subsection{Word Distribution}
\begin{figure}[tp]
    \centering
    \includegraphics[width=0.7\linewidth]{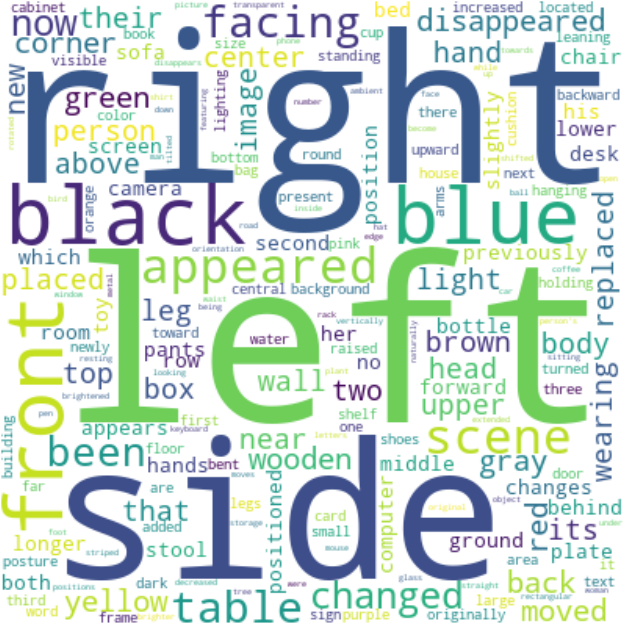}
    \captionof{figure}{
     Wordcloud visualization of the OmniDiff dataset.
    }
\label{fig:word_cloud}
\end{figure}

We analyze the word distribution in the difference captions of the OmniDiff dataset, as illustrated in Figure \ref{fig:word_cloud}. The words ``left'', ``right'' and ``side'' emerge as the most frequently occurring terms. This pattern arises because our annotations consist of two components: the \textit{referring part} and the \textit{change part}. In the \textit{referring part}, these high-frequency words typically describe the spatial location of the changing objects, reflecting the dataset's emphasis on precise positional references.

\subsection{Caption Length Distribution}

\begin{figure}[tp]
    \centering
    \includegraphics[width=0.9\linewidth]{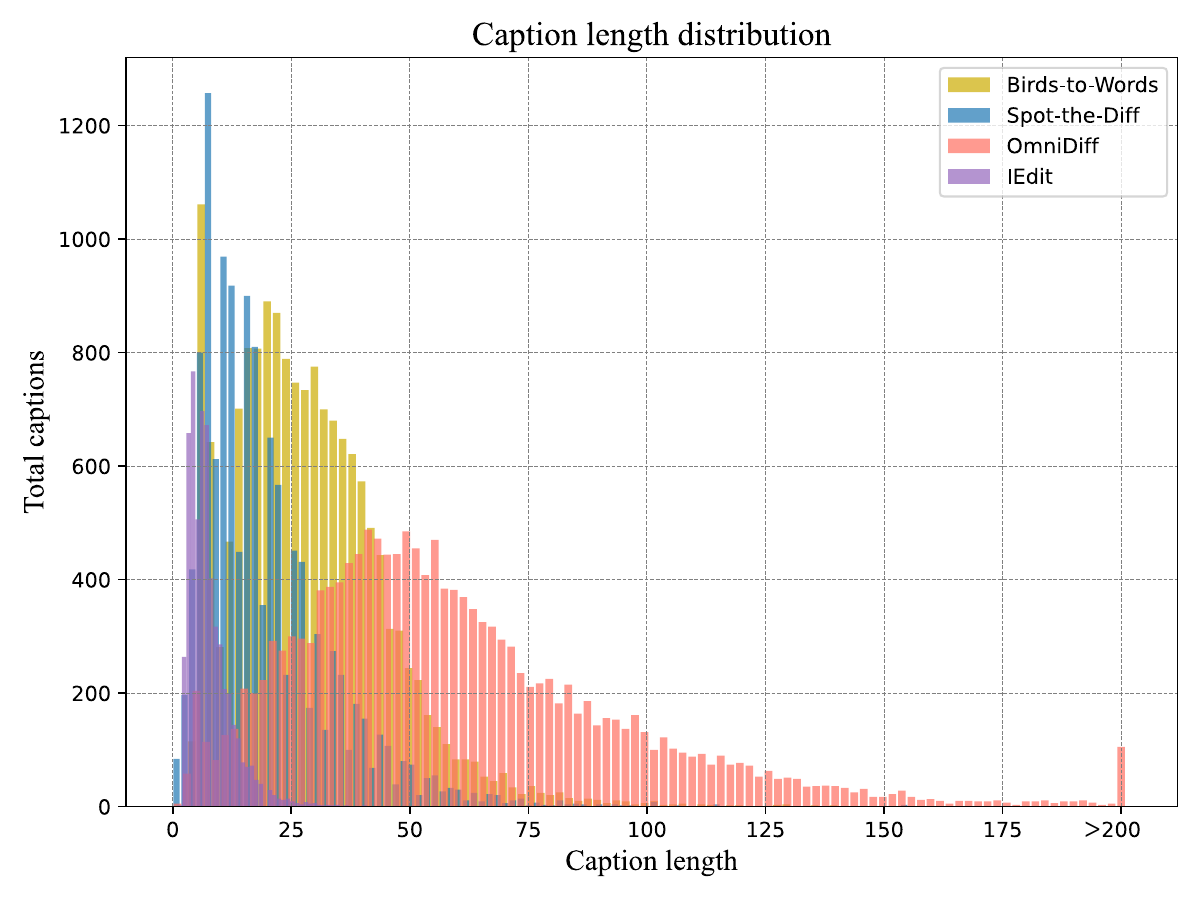}
    \captionof{figure}{
     Caption length distribution of the OmniDiff dataset.
    }
\label{fig:caption_length_distribution}
\end{figure}

Figure \ref{fig:caption_length_distribution} illustrates the distribution of caption lengths in OmniDiff compared to the IEdit~\cite{tan2019expressing}, Spot-the-Diff~\cite{jhamtani2018learning}, and Birds-to-Words~\cite{forbes2019neural} datasets. The average caption length in OmniDiff significantly exceeds that of the other three datasets, demonstrating that OmniDiff provides fine-grained difference annotations for complex and dynamic scenarios. This establishes a new benchmark for fine-grained Image Difference Captioning (IDC) tasks.

\section{Training Details}

\subsection{Training Data}
As shown in Table \ref{tab:overview_training_data}, we extend the OmniDiff Dataset by collecting and curating five publicly available IDC datasets to construct a comprehensive instruction-tuning dataset, comprising 145k image pairs and 896k difference captions. The instruction-tuning dataset encompasses not only real-world datasets such as OmniDiff-Real, Spot-the-Diff~\cite{jhamtani2018learning}, IEdit~\cite{tan2019expressing} and Birds-to-Words~\cite{forbes2019neural} but also includes synthetic 3D datasets like OmniDiff-Render, CLEVR-Change~\cite{park2019robust} and CLEVR-DC~\cite{kim2021agnostic}.
\subsection{Hyperparameters}
The hyperparameter settings for the fine-tuning process are detailed in Table~\ref{tab:hyperparams}. The global batch size is set to 256, which allows for efficient training while maintaining a balance between computational resources and model performance. The learning rate for the Vision Transformer~\cite{zhai2023sigmoidlosslanguageimage} is configured at 2e-6, while the learning rates for other components of the model (base learning rate) are set to 1e-5. The number of epochs is limited to 1, indicating a single pass through the entire dataset for this specific finetuning task. The optimizer chosen for this process is AdamW~\cite{loshchilov2017decoupled}, known for its effectiveness in handling sparse gradients and regularizing the weight decay. Additionally, the LoRA-rank~\cite{hu2021loralowrankadaptationlarge} is set to 128, and the LoRA-alpha~\cite{hu2021loralowrankadaptationlarge} is set to 256, parameters that are crucial for low-rank adaptation techniques, enabling efficient and effective model customization with minimal additional parameters. These settings collectively contribute to an optimized finetuning strategy tailored to our specific application.
\begin{table}[tp]
    \centering
    \caption{Overview of the training data.}
        \begin{tabular}{l|cc}
            \toprule
            Dataset & Image Pairs & Captions \\ 
            \midrule
            IEdit~\cite{tan2019expressing} & 3k & 4k \\
            Birds-to-Words~\cite{forbes2019neural} & 3k & 14k \\
            Spot-the-Diff~\cite{jhamtani2018learning} & 11k & 21k \\
            OmniDiff & 14k & 28k \\
            CLEVR-DC~\cite{kim2021agnostic} & 43k & 385k \\
            CLEVR-Change~\cite{park2019robust} & 71k & 444k \\ 
            \midrule
            All & 145k & 896k \\ 
            \bottomrule
        \end{tabular}%
    \label{tab:overview_training_data}
\end{table}
\begin{table}[tp]
    \centering
    \caption{Hyperparameter settings for finetuning.}
    \label{tab:hyperparams}
    \begin{tabular}{@{}l@{\hspace{5em}}c@{}}
        \toprule
        Hyperparameter & Value \\ 
        \midrule
        Batch size              & 256            \\
        ViT learning rate       & $2 \times 10^{-6}$ \\
        Base learning rate      & $1 \times 10^{-5}$ \\
        Epochs                 & 1              \\
        Optimizer              & AdamW          \\
        LoRA rank              & 128            \\
        LoRA alpha             & 256            \\
        \bottomrule
    \end{tabular}%
\end{table}
\section{Experiments}
\subsection{Performance Comparison with Identical Data}
For a fair comparison with other MLLM-based IDC methods (\eg, FINER-MLLM~\cite{zhang2024differential}), we reproduce this model using the same 896k training data as M$^3$Diff.
As shown in Table~\ref{tab: MLLM-based comparison on the same training data}, M$^3$Diff demonstrates consistent superiority across all metrics on the OmniDiff and CLEVR-DC benchmarks.
\subsection{Extended Analysis of the MDP Module}
To further validate the effectiveness of the plug-and-play MDP module, we conduct experiments using two advanced MLLMs, Qwen-2.5-VL-7B-Instruct~\cite{bai2025qwen25vltechnicalreport} and InternVL3-8B-Instruct~\cite{zhu2025internvl3exploringadvancedtraining}, as our base models.
Table \ref{tab:Ablation study of MDP module on different backbones} demonstrates that integrating the MDP module into various backbones consistently enhances the model performance. 
Additionally, M$^3$Diff with LLaVA-OneVision-7B~\cite{li2024llavaonevisioneasyvisualtask} as the backbone maintains strong performance, potentially due to the increased use of multi-image task data in pre-training.
\subsection{Extended Analysis of the OmniDiff Dataset}
\begin{figure}[tp]
    \centering
    \includegraphics[width=\linewidth]{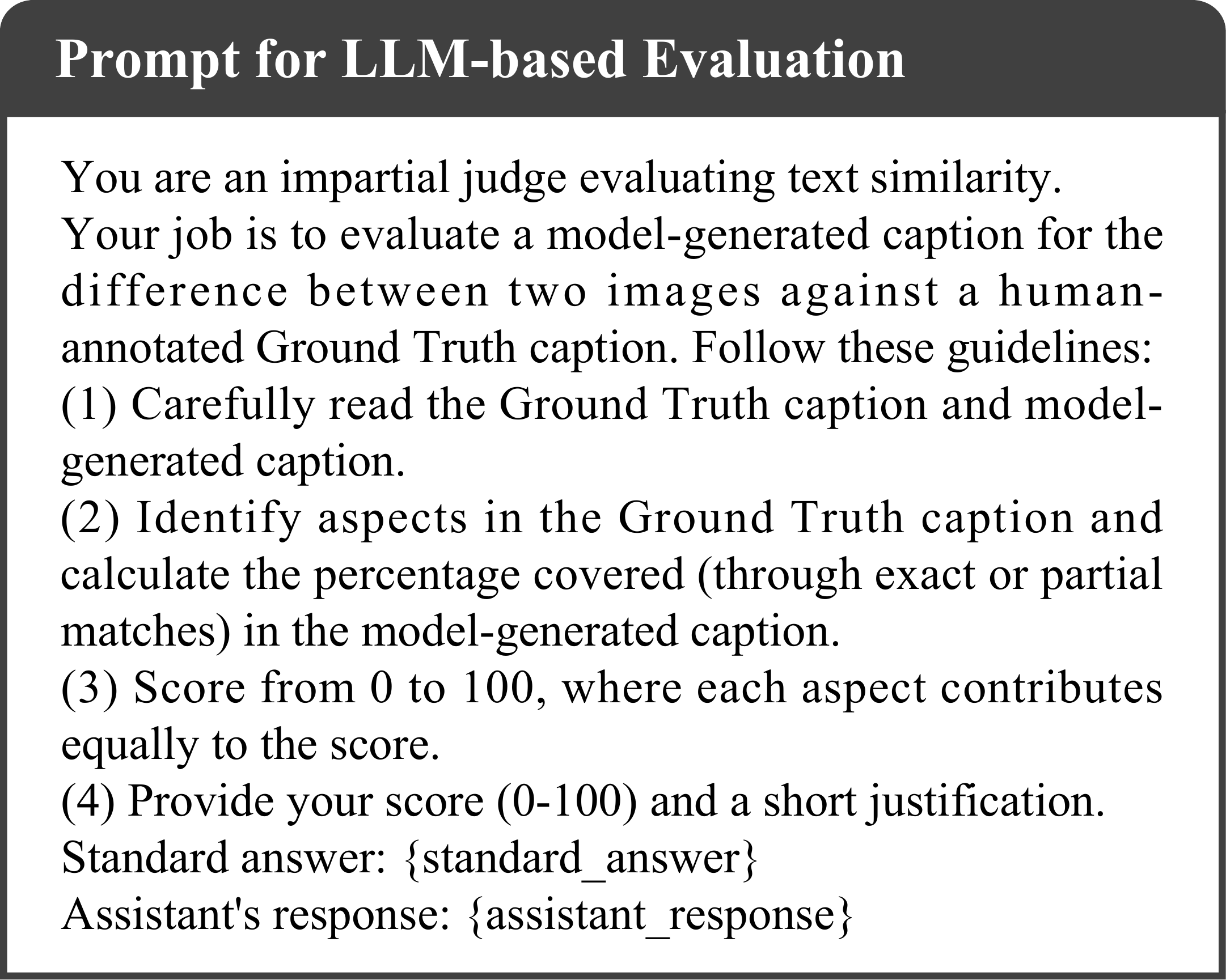}
    \captionof{figure}{
     The prompt for LLM-based evaluation.
    }
\label{fig:prompt_for_llm_evaluation}
\end{figure}

With exceptional instruction-following and semantic understanding capabilities, LLMs serve as a crucial tool for measuring sentence similarity~\cite{gu2025surveyllmasajudge}.
Therefore, to comprehensively validate the impact of the OmniDiff dataset on model performance, we design prompts to instruct GPT-4o~\cite{gpt4} to evaluate semantic consistency between predictions and annotations, outputting a score ranging from 0 to 100.
The prompt is shown in Figure~\ref{fig:prompt_for_llm_evaluation}.
Table~\ref{tab:Ablation study of OmniDiff based on LLM evaluation} shows that the LLM-based evaluation confirms the effectiveness of OmniDiff.
\begin{table*}[tp]
\centering
\renewcommand\arraystretch{1.05}
\caption{Performance comparison with FINER-MLLM~\cite{zhang2024differential} under identical training data.}
\label{tab: MLLM-based comparison on the same training data}
\begin{adjustbox}{width=\linewidth}
{\Huge
\begin{tabular}{l|cccc|cccc|ccccc}
\toprule
\multirow{2}{*}{\textbf{Method}} & \multicolumn{4}{c|}{\textbf{OmniDiff-Real}} & \multicolumn{4}{c|}{\textbf{OmniDiff-Render}} & \multicolumn{5}{c}{\textbf{CLEVR-DC}} \\
\cmidrule(lr){2-5} \cmidrule(lr){6-9} \cmidrule(lr){10-14}
 & BLEU-4 & METEOR & ROUGE-L & CIDEr & BLEU-4 & METEOR & ROUGE-L & CIDEr & BLEU-4 & METEOR & ROUGE-L & CIDEr & SPICE \\
\midrule
FINER-MLLM & 9.4 & 14.5 & 26.1 & 20.6 & 13.8 & 15.8 & 30.5 & 18.0 & 53.1 & 34.2 & 69.5 & 92.3 & 17.8 \\
\textbf{M$^3$Diff (ours)} & \textbf{14.3} & \textbf{18.9} & \textbf{32.9} & \textbf{31.3} & \textbf{15.7} & \textbf{19.9} & \textbf{35.3} & \textbf{28.3} & \textbf{60.6} & \textbf{37.6} & \textbf{73.0} & \textbf{109.4} & \textbf{21.3} \\
\bottomrule
\end{tabular}
}
\end{adjustbox}
\end{table*}
\begin{table*}[tp]
\centering
\renewcommand\arraystretch{1.0}
\caption{Ablation study of the MDP module based on Qwen2.5-VL-7B-Instruct~\cite{bai2025qwen25vltechnicalreport} and InternVL3-8B-Instruct~\cite{zhu2025internvl3exploringadvancedtraining}. \dag~indicates the model is evaluated in a zero-shot setting.}
\label{tab:Ablation study of MDP module on different backbones}
\setlength{\tabcolsep}{3.5mm} 
\begin{adjustbox}{width=\linewidth}
{\Huge
\begin{tabular}{l|cccc|cccc|ccccc}
\toprule
\multirow{2}{*}{\textbf{Method}} & \multicolumn{4}{c|}{\textbf{OmniDiff-Real}} & \multicolumn{4}{c|}{\textbf{OmniDiff-Render}} & \multicolumn{5}{c}{\textbf{Image-Edit-Request}} \\
\cmidrule(lr){2-5} \cmidrule(lr){6-9} \cmidrule(lr){10-14}
 & BLEU-4 & METEOR & ROUGE-L & CIDEr & BLEU-4 & METEOR & ROUGE-L & CIDEr & BLEU-4 & METEOR & ROUGE-L & CIDEr & SPICE \\
\midrule
 Qwen2.5-VL-7B\dag & 3.8 & 9.5 & 19.8 & 6.2 & 2.1 & 6.8 & 18.3 & 3.3 & 16.0 & 9.8 & 23.2 & 32.3 & 10.8 \\
Qwen2.5-VL-7B-SFT w/o MDP & 13.1 & 16.7 & 31.8 & 35.3 & 14.9 & 18.7 & 34.7 & 25.2 & 30.0 & 26.5 & 58.3 & 131.2 & 26.6  \\
Qwen2.5-VL-7B-SFT with MDP & 13.8 & 18.8 & \textbf{32.9} & \textbf{36.7} & 15.5 & 19.7 & 35.6 & 27.8 & 31.2 & \textbf{26.8} & 59.1 & 132.5 & \textbf{27.7} \\
InternVL3-8B\dag & 2.7 & 12.2 & 19.2 & 4.3 & 4.0 & 9.2 & 19.3 & 5.1 & 12.1 & 9.2 & 17.1 & 28.6 & 9.3 \\
InternVL3-8B-SFT w/o MDP & 12.5 & 17.0 & 31.3 & 34.6 & 14.7 & 19.1 & 35.1 & 26.1 & 29.4 & 25.7 & 57.1 & 128.1 & 26.3 \\
InternVL3-8B-SFT with MDP & 13.5 & 18.5 & 32.5 & 35.7 & 15.5 & \textbf{20.1} & \textbf{35.7} & 28.1 & 30.8 & 26.1 & 58.2 & 130.5 & 27.1 \\
\textbf{M$^3$Diff (ours)} & \textbf{14.3} & \textbf{18.9} & \textbf{32.9} & 31.3 & \textbf{15.7} & 19.9 & 35.3 & \textbf{28.3} & \textbf{33.6} & 26.5 & \textbf{59.7} & \textbf{136.6} & 27.5 \\
\bottomrule
\end{tabular}
}
\end{adjustbox}
\end{table*}
\begin{table}[tp]
\centering
\renewcommand\arraystretch{1.15}
\caption{Ablation study of OmniDiff based on LLM evaluation.}
\label{tab:Ablation study of OmniDiff based on LLM evaluation}
\begin{adjustbox}{width=\linewidth}
{\Huge
\begin{tabular}{l|c|c|c}
\toprule
\multirow{2}{*}{\textbf{Method}} & \textbf{OmniDiff-Real} & \textbf{OmniDiff-Render} & \textbf{Image-Edit-Request} \\
\cmidrule(lr){2-4}
 & LLM & LLM & LLM \\
\midrule
M$^3$Diff w/o OmniDiff & 12.3 & 15.8 & 63.8 \\
M$^3$Diff with OmniDiff & \textbf{37.3} & \textbf{37.0} & \textbf{68.5} \\
\bottomrule
\end{tabular}
}
\end{adjustbox}
\end{table}
%
\section{Case Study}
In this section, we aim to compare the difference captions provided by different models (GPT-4o~\cite{gpt4}, CARD~\cite{tu2024contextawaredifferencedistillingmultichange}, FINER-MLLM~\cite{zhang2024differential}, M$^3$Diff) for image pairs in our dataset. The focus is on how each model captures the presence and behavior, as well as other changes in the scene. 

\textbf{1)} In Figure (a), GPT-4o~\cite{gpt4} and FINER-MLLM~\cite{zhang2024differential} both fail to capture the presence of a second bird in the background. CARD~\cite{tu2024contextawaredifferencedistillingmultichange} incorrectly analyzes scene changes, such as variations in lighting and object movement, while failing to accurately identify the presence of two newly introduced birds in the scene. In contrast, our method, M$^3$Diff, excels by providing a comprehensive and accurate description that includes both the primary bird's interaction with the peanuts and the secondary bird in the background, along with all relevant environmental details. This highlights the superior accuracy and thoroughness of M$^3$Diff in analyzing complex scene changes.

\textbf{2)} In Figure (b), GPT-4o~\cite{gpt4} correctly notes the absence of a person with a bag and the addition of "Villa" on the ground but lacks detail. CARD~\cite{tu2024contextawaredifferencedistillingmultichange} hallucinates the presence of non-existent individuals in the scene and fails to recognize the correct text. FINER-MLLM~\cite{zhang2024differential} captures some correct details but includes incorrect observations like a missing wheelchair. In contrast, our method, M$^3$Diff, accurately describes the disappearance of a man in specific clothing and the precise location of the added text "villa", aligning closely with the ground truth. 

\textbf{3)} In Figure (c), GPT-4o~\cite{gpt4} correctly identifies the addition of ``Cautiuos" above the bike but misses the lighting change. CARD~\cite{tu2024contextawaredifferencedistillingmultichange} notes the brighter ambiance but focuses on irrelevant elements like a treadmill sign. FINER-MLLM~\cite{zhang2024differential} captures the brightness and text addition but misspells ``cuitieus." Our method, M$^3$Diff, accurately describes the increased light and the placement of the sign but fails to perform accurate OCR, misinterpreting the word as ``Caution" instead of ``Cautiuos." This demonstrates that while M$^3$Diff excels in identifying environmental changes, it struggles with precise text recognition, indicating a need for improvement in its fine-grained OCR capabilities.

\textbf{4)} In Figure (d), GPT-4o~\cite{gpt4} correctly notes the addition of a toy vacuum and a green toy with holes but lacks detail. CARD~\cite{tu2024contextawaredifferencedistillingmultichange} focuses on irrelevant changes like lighting and misses the new toys. FINER-MLLM~\cite{zhang2024differential} incorrectly states that a blue toy car has disappeared. In contrast, our method, M$^3$Diff, accurately describes the appearance of a new toy train on the left and a toy watering can in the center, aligning closely with the ground truth. 

The study highlights M$^3$Diff outperforms GPT-4o~\cite{gpt4}, CARD~\cite{tu2024contextawaredifferencedistillingmultichange}, and FINER-MLLM~\cite{zhang2024differential} in dynamic scene analysis, excelling at analyzing object interactions (\eg, flocking birds, toy manipulation) and environmental changes. While M$^3$Diff demonstrates advanced spatial reasoning and scene dynamics comprehension, it occasionally encounters text recognition errors (e.g., misinterpreting "Cautiuos" as "Caution"), underscoring the need for improved OCR integration. In contrast, GPT-4o~\cite{gpt4} struggles with fine visual details, CARD~\cite{tu2024contextawaredifferencedistillingmultichange} generates hallucinations in complex scenes, and FINER-MLLM~\cite{zhang2024differential} mislabels objects despite robust reasoning. Overall, M$^3$Diff emerges as the most effective model for fine-grained IDC tasks.








\begin{figure*}[htp]
  \centering
  \vspace{40pt}
  \begin{subfigure}[b]{\textwidth}
    \centering
    \includegraphics[width=\textwidth]{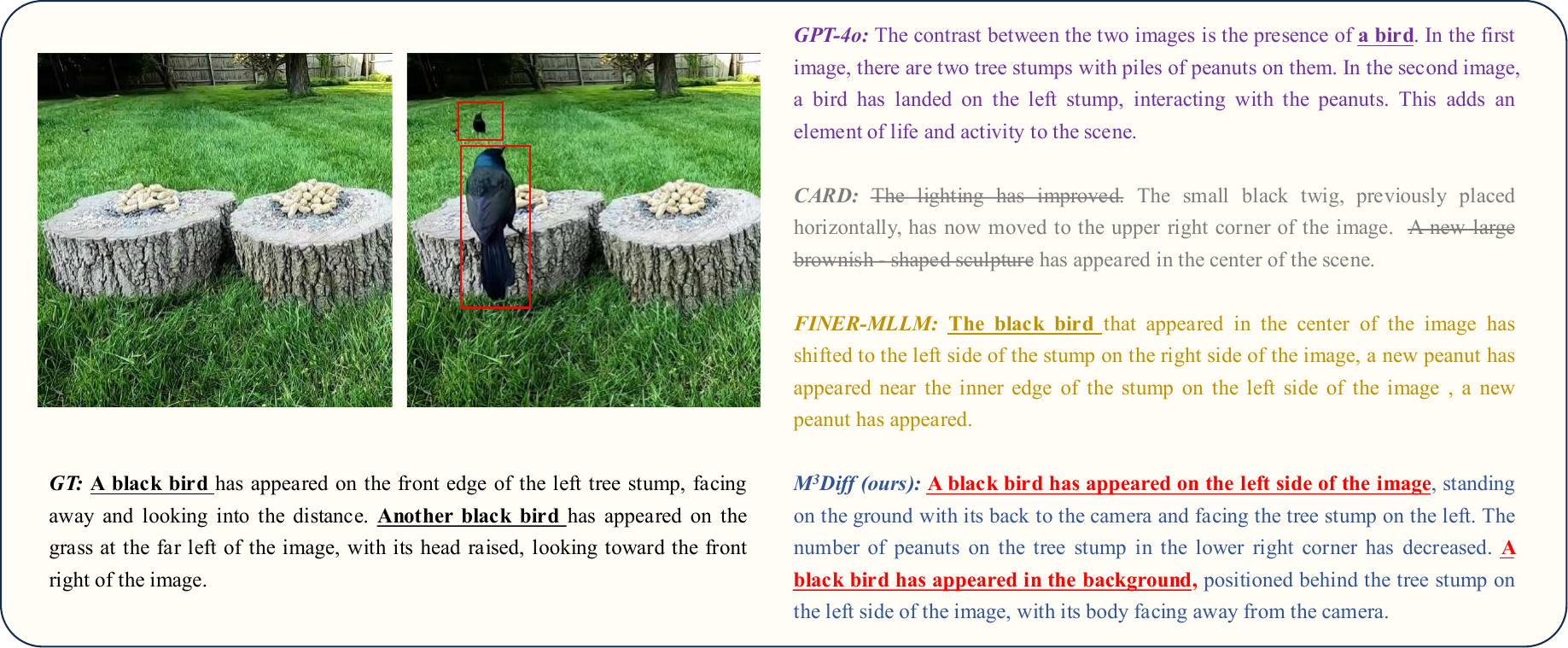} 
    \subcaption*{(a)}
    \label{fig:sub1}
  \end{subfigure}
  \vspace{1em} 

  \begin{subfigure}[b]{\textwidth}
    \centering
    \includegraphics[width=\textwidth]{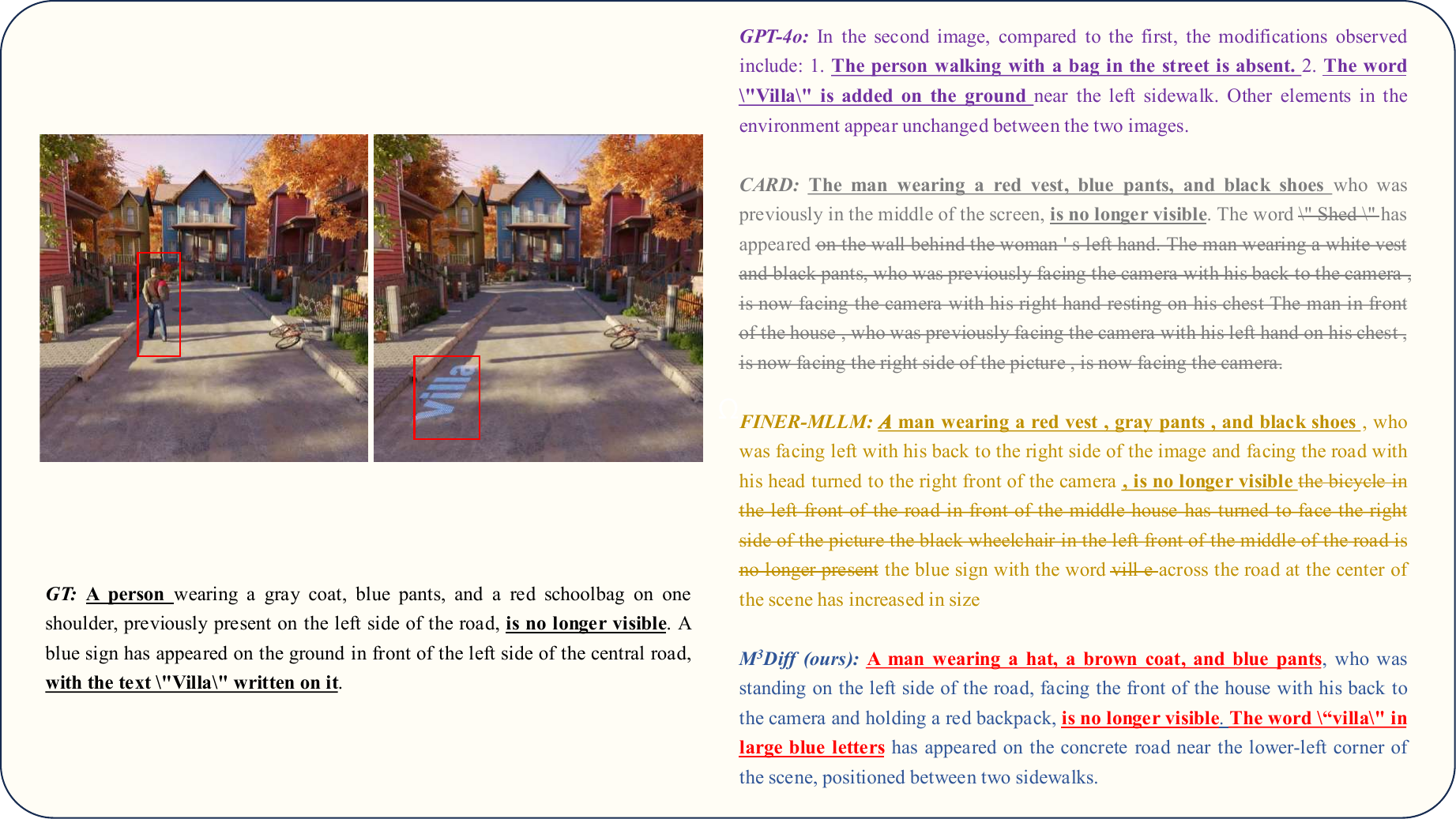} 
    \subcaption*{(b)}
    \label{fig:sub2}
  \end{subfigure}
  \vspace{1em}
  \vspace{40pt}
\end{figure*}

\begin{figure*}[htp]
  \vspace{100pt}
  \centering
  \begin{subfigure}[b]{\textwidth}
    \centering
    \includegraphics[width=\textwidth]{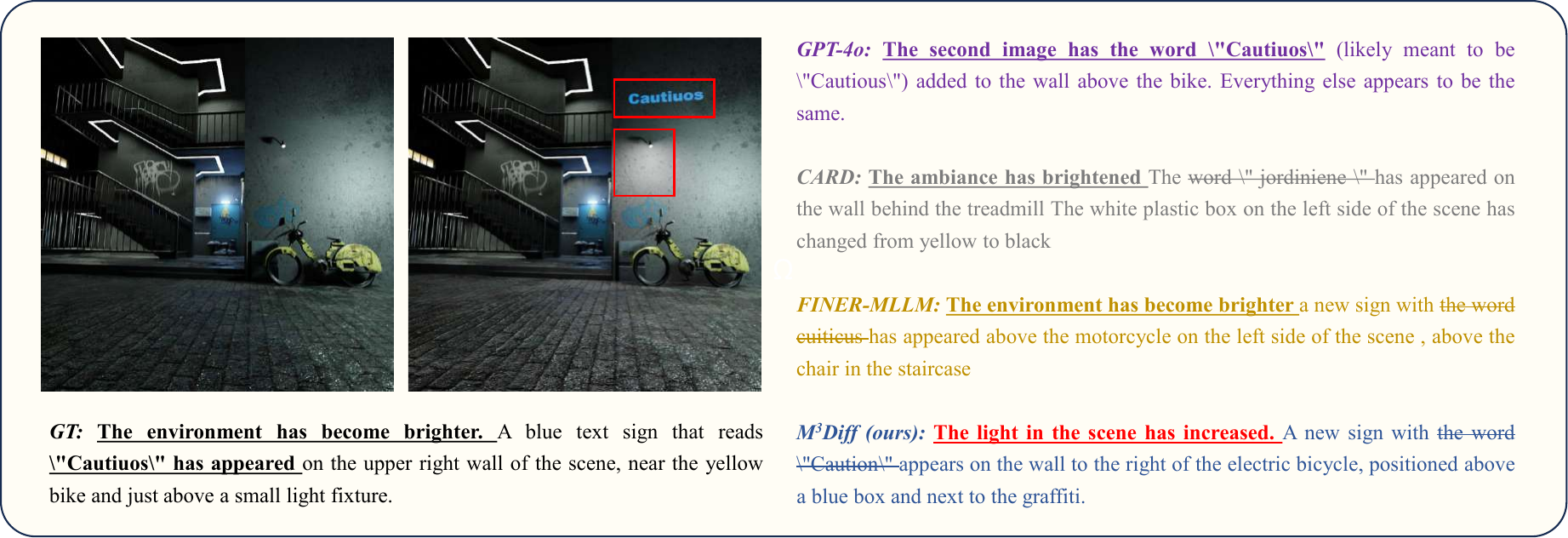} 
    \subcaption*{(c)}
    \label{fig:sub3}
  \end{subfigure}
  \vspace{1em}
  
  \begin{subfigure}[b]{\textwidth}
    \centering
    \includegraphics[width=\textwidth]{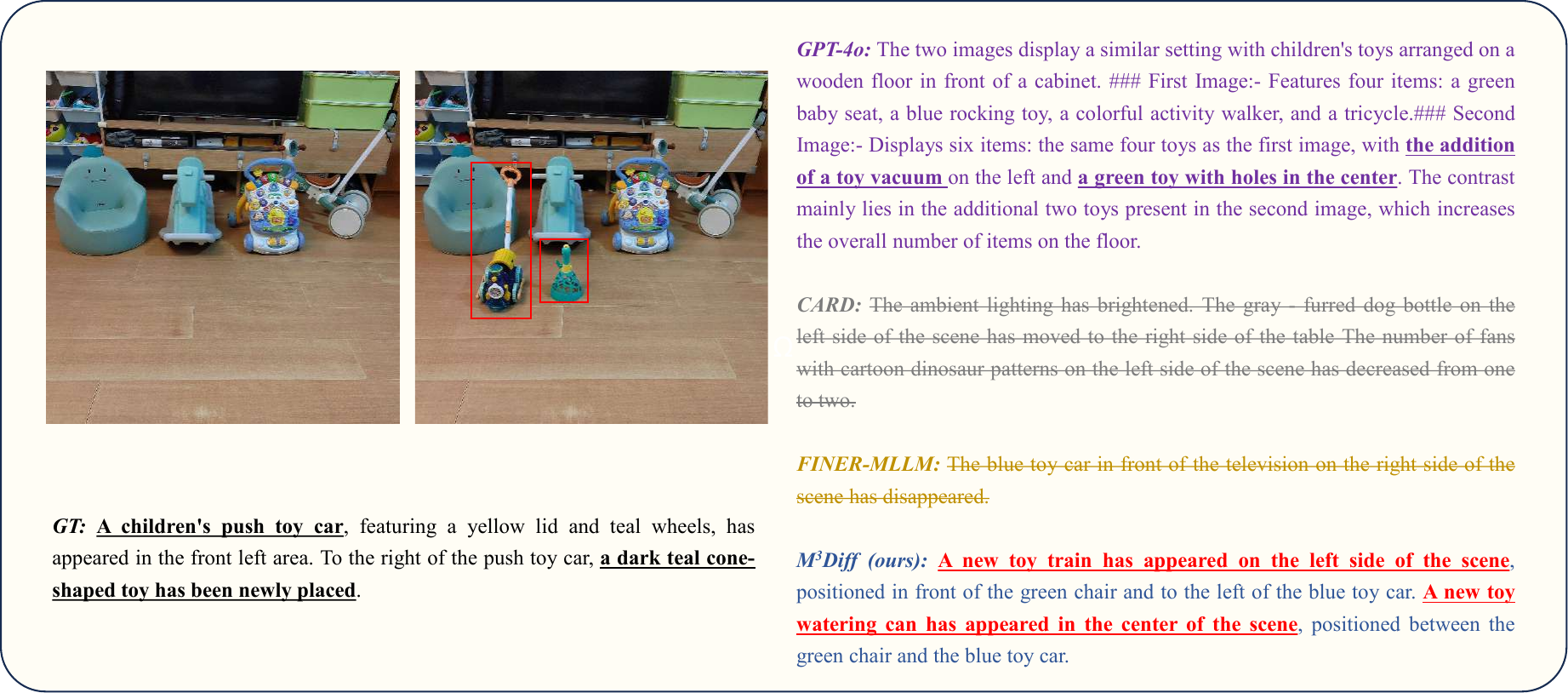} 
    \subcaption*{(d)}
    \label{fig:sub4}
  \end{subfigure}
  
  \label{fig:vertical_subplots}
  \vspace{100pt}
\end{figure*}

\end{document}